\documentclass[10pt,table]{cai}

\usepackage{booktabs}
\usepackage{graphicx}
\usepackage{hyperref}
\usepackage{expex}

\usepackage[frozencache,cachedir=.]{minted} 
\usepackage{gb4e} 
\graphicspath{{figs/}}

\begin{document}

\volumeheader{36}{0}%{00.000}
\begin{center}

  \title{Radar de Parité: An NLP system to measure gender representation in French news stories}
  \maketitle

  \thispagestyle{empty}

  \begin{tabular}{cc}
      Valentin-Gabriel Soumah\upstairs{\affilone}, Prashanth Rao\upstairs{\affilone}, Philipp Eibl\upstairs{\affilone}, Maite Taboada\upstairs{\affilone,*}
  \\[0.25ex]
    {\small \upstairs{\affilone}Discourse Processing Lab, Simon Fraser University, Canada} \\
  \end{tabular}

   \emails{\upstairs{*}Corresponding author: mtaboada@sfu.ca}
\end{center}

\begin{abstract}
\noindent We present the Radar de Parité, an automated Natural Language Processing (NLP) system that measures the proportion of women and men quoted daily in six Canadian French-language media outlets. 
%The Radar follows our work on the English-language Gender Gap Tracker. 
We outline the system's architecture and detail the challenges we overcame to address French-specific issues, in particular regarding coreference resolution, a new contribution to the NLP literature on French. We also showcase statistics covering over one year's worth of data (282,512 news articles). Our results highlight the underrepresentation of women in news stories, while also illustrating the application of modern NLP methods to measure gender representation and address societal issues.
\end{abstract}

\begin{keywords}{Keywords:}
Natural language processing, French, quote extraction, coreference, news, gender
\end{keywords}
\copyrightnotice

\vspace{-0.2cm}
\section{Gender representation in news stories}
\label{sec:intro}
\vspace{-0.2cm}

Natural Language Processing (NLP) has excellent potential for applied research in various areas, such as capturing sentiment in reviews or news stories, highlighting toxicity in social media, or mining the biomedical literature. The commonality in most applied NLP research projects is the need to reliably and scalably extract information from unstructured text data. In this paper, we describe one such application: extracting quotes from news stories to quantify gender representation.

Gender representation in the media is a long debated topic. From the 1970s, there have been studies into how much women and gender-diverse people are portrayed in news stories, with the general hypothesis that they tend to be underrepresented \cite{Macharia20-WMT, Kassova20-TMP2}. There is also research studying \textit{how} they are represented, i.e., whether sexist or homophobic tropes are present when we discuss women and gender-diverse people \cite{VanderPas20-GDI,Trimble20-GNA}. In this work, we tackle one specific aspect of representation: who is quoted and in what proportions. Our starting hypothesis is that we hear less from women than from men in news stories, that is, that men are quoted more often than is to be expected from their proportion in the general population. To fully answer this question, we formulate a quantitative approach, collecting large amounts of representative data and extracting quotes from the unstructured text. This is the goal of the \textit{Radar de Parité}. 

We define quotes as either direct or indirect reproductions of what a person said, and we define that person as a \textit{source} in news articles. In order to extract quotes, we employ a full NLP pipeline, focusing on parsing to identify speakers, verbs, and quotes, in each news story. We then predict the gender of the speaker (or source), using external gender-prediction services. Finally, we display the results on a public-facing dashboard.\footnote{French dashboard: \url{https://radardeparite.femmesexpertes.ca/} \\ \hspace*{0.45em}English dashboard: \url{https://gendergaptracker.informedopinions.org/} \\ \hspace*{0.45em}Code: \url{https://github.com/sfu-discourse-lab/GenderGapTracker}}

The contributions of our paper are multifold. First, we address the difficulty of doing large-scale NLP in languages other than English. We show that, even for a relatively well-resourced language such as French, off-the-shelf language models and specialized NLP modules are not always readily available. Next, we show how we built a modular system for applied research, where the goal is not to improve a foundational algorithm or to achieve state of the art results, but rather to produce a reliable, accurate system for a specific use case. Finally, we present gender statistics of the Radar de Parité on over a year's worth of data, and show that women are, indeed, underrepresented in mainstream media in Canada. 

\vspace{-0.2cm}
\section{Multilingual NLP and the hegemony of English}
\label{sec:multilingual}
\vspace{-0.2cm}

One of the promises of NLP is the ability to reuse resources from one language and apply them to another. In principle, many innovations in tokenization, part-of-speech tagging, parsing, word sense disambiguation, and corerefence resolution, if developed for English, can carry over to other languages, after adjusting for linguistic differences. In practice, however, ready-to-use algorithms suffer from a lack of annotated data to train on.
This is the case even after the advent of large language models, most of them trained on English-language data \cite{ruder2022statemultilingualai,Bommasani21-OTO}. Chinese is perhaps the exception \cite{zeng2022glm130b}, but we welcome the recent efforts of groups such as HuggingFace in providing large open-source multilingual models \cite{Scao22-BLO}. We see this lack of resources for languages other than English as an obstacle in the practical application of NLP. Despite the inherent power of transformer-based language models, as well as related breakthroughs in multi-modal and transfer learning, their computing requirements are still a challenge when applied at scale \cite{ruder2022statemultilingualai}.

As a result, we rely on \texttt{spaCy}, an industrial-strength NLP library (see Section \ref{sec:pipeline}) to perform the analysis shown in this paper. We chose \texttt{spaCy} because it is robust, well supported, and regularly updated, with a full pipeline. There are perhaps state-of-the-art resources for each of the components (e.g., for French NER \cite{Choudhry23-TBN}), but integration into a pipeline is at best complex and often impossible.
In prior work, we had built a quote detection system for English, which is well supported by \texttt{spaCy}'s existing English language models \cite{Asr20-GGT,Rao21-GBI}. For French, we lacked a module for coreference resolution. \texttt{spaCy}, until recently, provided it via \texttt{neuralcoref} \cite{Clark16-DRL}, a pre-trained and trainable neural coreference module.\footnote{\url{https://spacy.io/universe/project/neuralcoref}} \texttt{neuralcoref}, however, works only in English, with no equivalent functionality for other languages. As we detail in the next section, we ended up adopting and expanding another library, \texttt{coreferee}, which supports English, French, German, and Polish.\footnote{\url{https://spacy.io/universe/project/coreferee}}

\vspace{-0.2cm}
\section{NLP pipeline}
\label{sec:pipeline}
\vspace{-0.2cm}

\begin{figure}[ht]
    \centering
    \includegraphics[scale=0.8]{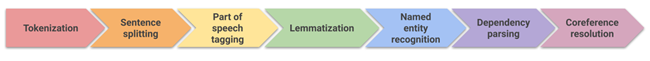} \\
    \caption{NLP pipeline}
    \label{fig:pipeline}
\end{figure}

We scrape news articles daily from six French-language outlets in Canada. We store the articles and their metadata to a database and pass it to our NLP pipeline. We apply a typical sequence of steps to preprocess unstructured text data, including tokenization and part-of-speech tagging, in order to extract information for downstream analysis (Figure \ref{fig:pipeline}).

The goal of Named Entity Recognition is to identify the names of people mentioned, a subset of which are typically quoted in a news article. Dependency parsing is used to extract syntactic quotes. Coreference resolution is a key step that allows us to cluster references to the same person (e.g., \textit{Valérie Plante, Mme Plante, she, her}).

\vspace{-0.2cm}
\subsection{Quote extraction}
\label{sec:quote-extraction}
\vspace{-0.2cm}
To measure the gender representation gap in news media, we identify the number of men and women who are quoted in news articles; in other words, people who have not only been mentioned, but have also seen their voices reflected in the news. We consider both direct speech, surrounded by quotation marks (\textit{She stated, ``\ldots''}), and indirect speech (\textit{She stated that\ldots}) to be quotations. We extract all the quotes from the news article text, following which we align the quoted speakers with the unified named entities that were gender-labelled (see Section \ref{sec:gender-prediction}). Based on our study of the literature on reported speech \cite{Coulmas86-RSS,Krestel08-MTS,Spronck19-RSF}, we separate quotes into three different types: 
direct quotes, indirect quotes, and \textit{selon} quotes. Examples of each are shown in Example (\ref{ex:quotes}). More detailed descriptions of the quote extraction process can be found in Appendix \ref{appendix:quote-extraction}.

%https://www.lapresse.ca/actualites/grand-montreal/201901/31/01-5212974-un-homme-lie-a-la-mafia-serait-un-proprietaire-de-la-grande-roue.php

\begin{small}
\begin{exe} 
\ex\label{ex:quotes}\textbf{Direct}: «N'entre pas qui veut dans le cercle de Vito Rizzuto. Importer 2000 lb de haschich à 26 ans, il n'a sûrement pas fait ça tout seul», dit M. Mansueto. \\
 \textbf{Indirect}: Le sergent-détective a également expliqué que les criminels avaient besoin de s'impliquer dans des activités légales pour blanchir leur argent sale. \\
 \textbf{With \textit{selon:}} Selon eux, l'individu n'offre qu'une partie des services de distribution alimentaire et n'a aucun intérêt dans la Grande Roue. 
\end{exe}
\end{small}
\vspace{-0.2cm}

\vspace{-0.2cm}
\subsection{Recognizing people}
\label{sec:ner}
\vspace{-0.2cm}
By \textit{people} we mean all human entities or persons that are mentioned in the news text. The term \textit{sources} refers to the subset of people who are quoted within a news article. We use Named Entity Recognition (NER), a commonly used procedure in NLP, to identify the mentioned people in each article. Current NER techniques, largely statistical in nature and based on neural network architectures, work fairly well on our French data, although we observed that they are less reliable than the techniques in our prior work on English \cite{Asr20-GGT}. 
We believe that the extensive efforts we put in to improve the recognition of people's names may be useful for other researchers, and thus provide detailed information in Appendix \ref{appendix:ner}.

\vspace{-0.2cm}
\subsection{Coreference resolution}
\label{sec:coreference}
\vspace{-0.2cm}
A coreference resolution algorithm takes in text and returns clusters of text spans that refer to the same entity. For our use case, these text spans are either pronouns, partial or full names of people, or noun phrases that refer to someone without naming them, for example, \textit{la comptable} (`the [female] accountant'). To apply coreference resolution, we use the \texttt{coreferee} library, including a Rules Analyzer for French built  for this project. Processing the data via \texttt{coreferee} involves several steps, detailed in the subsections below.  
%The output of this system when applied to our text is a set of clusters populated by different text spans that may or may not appear in our list of named entities.

\vspace{-0.2cm}
\subsubsection{Coreference clusters}
\vspace{-0.2cm}

The output of coreference resolution is a set of indexed clusters of mentions. A `mention' is a list of tokens (typically one) that correspond to the syntactic heads of noun (or pronoun) phrases.
%(see Appendix \ref{appendix:ner}). 
\texttt{coreferee} produces coreference `chains' for each mention, and for each chain we create a coreference `cluster'. The cluster is an array of mentions’ character spans where each list corresponds to one mention, and the character spans consist of the beginning and end index of a sequence of characters, as shown in Example (\ref{ex:selon-nos}).

\vspace{-0.2cm}
\begin{small}
\begin{exe} 
\ex\label{ex:selon-nos}Selon nos informations, M. Legault a commencé à ressentir les premiers symptômes durant le trajet de Québec vers Montréal, jeudi, après la période de questions à l’Assemblée nationale. \guillemotleft Ce sont des symptômes apparentés à un rhume\guillemotright, affirme son directeur des relations médias, Manuel Dionne.    Un test rapide s’est révélé positif et Legault a annoncé sur Twitter en début de soirée qu’il se plaçait en isolement, même s’il assure qu’il se sent \guillemotleft bien\guillemotright. \\
\textbf{Coreference chains:} \\
\texttt{0: M.(4), son(46), Legault(64), il(74), se(75), se(86)} \\
\texttt{1: directeur(47), Manuel(52)} 
 %   `According to our information, Mr. Legault began to feel the first symptoms during the trip from Quebec to Montreal, Thursday, after question period at the National Assembly. \\
%    ``These are symptoms similar to a cold,'' says its director of media relations, Manuel Dionne.\\
  %  A quick test turned out to be positive and Legault announced on Twitter at the start of the evening that he was placed in isolation, even if he assures that he feels ``fine.''
\end{exe}
\end{small}
\vspace{-0.2cm}

\vspace{-0.4cm}
\subsubsection{Mention-entity alignment}
\vspace{-0.2cm}

Next, we align each of these clusters from the coreference algorithm to the named entities extracted in the named entity recognition step. To perform alignment, for each named entity and cluster pair, we search if the named entity has span coverage (i.e., overlaps) with all the heads of a mention belonging to the cluster. 
If so, we infer that the coreference cluster contains mentions of that named entity.
Targeting the heads and not the whole noun phrase allows us to exclude entities that overlap with the noun phrase, but not with the head, dealing with several alignment problems. Named entities that cannot be assigned to another existing mention are considered singletons. In the example below, \textit{Christine Boyle} and \textit{Jean Swanson} are mentioned only once, and are hence considered singletons.

\vspace{-0.2cm}
\begin{small}
\begin{exe}
\ex\label{ex:kennedy-stewart}Kennedy Stewart: \texttt{[Kennedy Stewart, Monsieur Kennedy Stewart, son]} \\
Christine Boyle: \texttt{[Christine Boyle]} \\
Jean Swanson: \texttt{[Jean Swanson]} \\
\end{exe}
\end{small}
\vspace{-0.4cm}

\vspace{-0.5cm}
\subsubsection{Entity unification}
\vspace{-0.2cm}

Once we align named entities to their mention clusters, we still need to merge some clusters, because the coreference resolution algorithm is not perfect---it does not directly combine all mentions of a named entity across a document, especially if the entity is mentioned in two places that are far from each other within the text. Here, we apply some domain knowledge about human names in order to find clusters pointing to the same person. 

First, we extract different potential components of a name: the first name, the middle name(s), and the last name. The middle names can correspond to one of the following: (a) cases where there is no clear last name (typically with names from Middle Eastern or Latin American origin, such as \textit{Mohammed bin Salman Al Saud}, or \textit{Andrés Manuel López Obrador}; the last part of the first name; (b) composite middle names (e.g., \textit{Ursula von der Leyen}); or (c) the spouse’s name.

Our name extraction system compares each part of the name individually between named entities representing different clusters. Several cases of matching are covered:

\begin{small}
\begin{itemize}
    \item Same last name and same first name: Due to computational cost, the coreference algorithm may not keep track of multiple mentions of a specific person in a long text. Sometimes, two full names that are exactly the same fall into separate clusters. We assume two full names that match exactly always refer to the same person, and combine their clusters.
    \item One shared name: One other common case is when some mentions in the text are only a first name or last name. For example, we can have three different entities aligned to three different clusters for one person: \textit{Justin Trudeau; Justin; Monsieur Trudeau; Trudeau}. In that case it's safe to assume they are the same person. There are possible cases of different people sharing one name but not the other. This typically happens with people of the same family, e.g., \textit{Sophie Grégoire Trudeau}.
\end{itemize}    
\end{small}

When either the first name or last name is shared but the other one is different, we infer that the two entities are different people. A possible exception is when two different people who share a last name are mentioned with only their last name. Those cases, however, are relatively rare, since journalists tend to use full names to avoid ambiguity in their writing.

When merging two clusters, we try to preserve the named entity mention that is deemed the most representative. This helps us prioritize the full-name representations for the cluster, which is useful for the gender prediction step. The priority order also uses the extracted name parts and is fairly straightforward:

\begin{small}
\begin{itemize}
    \item Presence of both first name and last name is more representative
    \item Presence of last name is more representative
    \item Presence of first name is more representative
    \item More middle names is more representative
\end{itemize}
\end{small}
    
After these steps, we come up with a unique cluster for each person containing all mentions in the text referring to that person, represented by one full name. We then move on to the next step of mapping the extracted quotes to the names of people who said them.

\vspace{-0.2cm}
\subsection{Mapping quote speakers to their references}
\label{sec:mapping-speakers}
\vspace{-0.2cm}

To identify the name and gender of the sources (quoted speakers), we find the corresponding named entity for each extracted quote (the reference), in three successive steps.

First, we compare the speaker index field of a quote to the indices of each named-entity-mention in our unified coreference clusters. If a mention span (rather, all its heads) and a speaker span have two or more overlapping characters, we assume the mention is the speaker and attribute the quote to the unified named entity (coreference cluster) of the mention. 

After trying to align all quotation speakers with potential named entities, there may still remain some quotes with speakers that could not be matched to named entities. We check if the text of the speaker matches with one of the text representatives of the clusters. If a match is found, the corresponding entity’s representative is assigned. This can happen when the named entity recognition step failed to identify a name somewhere in the text, but managed to identify the same name somewhere else (because of different contextual clues). 

As a last resort, we use a custom Rules Analyzer to check if the entity is a potential introducing common noun, i.e., a noun referring to a person that is not necessarily named anywhere in the text before. This is typically the case in nouns with an indefinite article (e.g., `a nurse'). In those cases, the text of the speaker is assigned as the reference. 
Failing all the methods above, our current system ignores the speaker and leaves the quote without a reference. There are several categories of these cases, such as quotes with a pronoun speaker (e.g., \textit{she said}) where the pronoun is still a singleton after all named entity and coreference cluster merging steps are finished. We provide statistics in Section \ref{sec:evaluation}. 

\vspace{-0.2cm}
\subsection{Gender prediction}
\label{sec:gender-prediction}
\vspace{-0.2cm}

We do acknowledge that gender is non-binary, that there are different social, cultural, and linguistic conceptualizations of gender, and that a binary is an oversimplification of the reality of gender in the population. For this project, however, we rely on self- and other-identification of gender through names and pronouns in order to classify people mentioned and quoted as women, men, or other. This is largely due to our use of external gender services that only offer binary classification of names.

Our approach to assigning gender to a first or full name was inspired by our prior work on English \cite{Asr20-GGT}, albeit with an extra step: title-based gender prediction, where we gain useful context from a potential title associated with the named entity (e.g., \textit{Monsieur Justin Trudeau}). This is because titles are more extensively gendered in French than in English. 

For gender prediction, %in cases where we have a person's full name, 
we utilize external gender services, which can be divided into two main groups: services that use only the first names (assigning gender based on statistics for those names from historical data) and services that rely on the full name. We use the services on a daily basis, and then sometimes correct errors manually as we encounter them. Genderize and Gender-API\footnote{\url{https://genderize.io/} and \url{https://gender-api.com/}} are two such services.

\vspace{-0.2cm}
\subsection{Gender annotation}
\label{sec:gender-annotation}
\vspace{-0.2cm}

Based on all the previous steps, we produce a structured object for each news article with fields in Table \ref{tab:fields}, with people-related fields on the top half and fields for each quote on the bottom (see also Figure \ref{fig:quote} in the Appendix for an example of the quote fields). 

%\vspace{-0.2cm}
\begin{small}
\begin{table}[ht]
\centering
\begin{tabular}{p{1in}p{4.2in}}
\midrule
{\cellcolor[HTML]{EFEFEF}\textbf{Field}} & {\cellcolor[HTML]{EFEFEF}\textbf{Description}}  \\ \midrule
People mentioned & All individual people mentioned in the text \\
Women mentioned & People from the list above of who were predicted to be female  \\ 
Men mentioned & People who were predicted to be male  \\
Other mentioned & People for whom gender prediction was unable to determine gender or who identify as non-binary \\
Sources & People who have at least one quote in the text. A subset of People mentioned. \\
Women sources & Sources who were predicted to be female  \\
Men sources & Sources who were predicted to be male  \\
Other sources & Sources for whom gender prediction was unable to determine gender or who identify as non-binary   \\
 \midrule
Speaker & Speaker of the quote + its start and end character indexes in the original text \\
Quote & Text of the quote + indexes \\
Verb & Quote indicating verb (if it exists) + indexes \\
Quote count & Length of the quote in words \\ 
Quote type & Direct, indirect, or \textit{selon} \\ 
Floating & Whether the quote is floating or not \\
  \bottomrule
\end{tabular}
\caption{People and quote fields produced by the NLP system}
\label{tab:fields}
\end{table}
\end{small}
%\vspace{-0.2cm}

% removed from table: 
% We only keep individual people by discarding the entities with conjunctions or commas. We also discard names that are less than two tokens long (to ensure we only have full names for gender prediction). 

%\vspace{-0.2cm}
\section{Evaluation}
\label{sec:evaluation}
\vspace{-0.2cm}

Evaluation was carried out separately for each component (people and source extraction, gender prediction, and quotation extraction) several times over the course of this work to test out new ideas that enhance the system. In this section, we explain our evaluation methodology and our current system's performance vs. human-annotated data.

\vspace{-0.2cm}
\subsection{Manual annotation for French articles}
\label{sec:manual-annotation}
\vspace{-0.2cm}

We prepared a collection of human-annotated gold samples to compare against our system's predicted quotes, people, and sources. We selected a sample of nine articles from each of the six newspapers' websites we scrape (\textit{Journal de Montréal, La Presse, Le Devoir, Le Droit, Radio-Canada, TVA Nouvelles}), for a total of 54 articles. 

We had one annotator who was fluent in French and English do the entire annotation. We began by using the English annotation guidelines, and extended the guidelines with French examples using feedback from the annotator, especially for cases that did not have parallels in English. For each article, we have a JSON file which contains an array of extracted quotes, verbs, and speakers, together with their character span index in the text (see Appendix \ref{appendix:quote-extraction}), where the speaker, quote, and verb are identified by their position in the file.
The annotation also added each person named in the articles as well as their gender. 

\vspace{-0.2cm}
\subsection{Quote evaluation}
\vspace{-0.2cm}

%To evaluate quote extraction performance, we annotated 98 articles published between December 2018 and February 2019. 
We evaluate the output of the system by comparing it to the human annotations. First, we align the annotations with the extracted quotes. Let $q_{a}$ be the span of annotated quote and $q_{e}$ be the span of the extracted quote. The match between the two quotes is defined as:

\begin{equation}
 score =   \frac{len(q_{a} \cap q_{e})}{len(q_{a})} 
\end{equation}

For each annotated quote $q_{a}$, the best matching quote in extracted quotes is the one with the highest matching score if the score is above a certain threshold (we experimented with 0.3 and 0.8 as easy and hard thresholds, respectively). In the following text example, the human annotated and automatically extracted quote spans are highlighted using italic and underlined text, respectively: ``\textit{It’s premature for us to make any sort of pronouncement about that right now, but \underline{I can tell you this thing looks and smells like a death penalty case}}.'' 

The alignment score is 0.45, which is the ratio of the length of the overlapping portion (69 characters) to the overall length of the annotated span (153 characters). Thus, this quote would be a match for the 0.3 threshold, but not for the 0.8 threshold. 

We evaluate the identification of speakers and verbs independently. We compute recall, precision and F1 score for the two, considering whether they are linked to the correct quote. That way, we can assess how many of the speakers will be identified for the next steps. %Since verbs and speakers are typically made of only one token, we use a relaxed match based on simple character overlap.

Table \ref{tab:quote-extraction} shows the result of evaluating the quotation extraction code on the manually-annotated dataset. The first three columns reflect how well the system captures the quotation content span (according to each of the set thresholds of overlap, 0.3 and 0.8) and the last two columns show system accuracy on verb and speaker detection. We consider the verb to have been correctly detected if the verb extracted by the system has exactly the same span as the expert-annotated span for the verb of that quotation. In order to evaluate the speaker detection quality at the surface textual level, we apply a simple overlap threshold: If the system-annotated span for the speaker has at least one character overlap with the expert-annotated text span for the speaker, it will be accepted as a correct annotation. 
For example, if the system-annotated span was \texttt{[12:25]}, corresponding to the string \textit{Valérie Plante}, while the human-annotated span was \texttt{[20:25]}, corresponding to the string \textit{Mme Plante}, the span overlap of 5 characters would mean they were considered the same speaker. Verb and speaker evaluations are applied only to the matched quotes (the quotations that are already passed as aligned between system and expert annotations based on the content span overlap). That is why the accuracy scores for Verb and Speaker in the table were higher when we used a stricter quote matching technique (hard match threshold).

\vspace{-0.2cm}
\begin{table}[ht]
\centering
\begin{tabular}{lccc|cc}
\midrule
                & \multicolumn{3}{c|}{\cellcolor[HTML]{EFEFEF}\textbf{Quotation content}}                                                  & \multicolumn{2}{c}{\cellcolor[HTML]{EFEFEF}\textbf{Speakers/verbs}}                                \\
                & \cellcolor[HTML]{EFEFEF}\textbf{Precision} & \cellcolor[HTML]{EFEFEF}\textbf{Recall} & \cellcolor[HTML]{EFEFEF}\textbf{F1-score} & \cellcolor[HTML]{EFEFEF}\textbf{Verb accuracy} & \cellcolor[HTML]{EFEFEF}\textbf{Speaker accuracy} \\ \midrule
\cellcolor[HTML]{EFEFEF}Easy match threshold (0.3) & 84.0\%                                     & 79.5\%                                  & 81.7\%                                    & 90.9\%                                         & 80.4\%                                            \\
\cellcolor[HTML]{EFEFEF}Hard match threshold (0.8) & 74.6\%                                     & 70.6\%                                  & 72.5\%                                    & 94.7\%                                         & 84.0\%                                            \\ \bottomrule
\end{tabular}
\caption{Quote extraction evaluation based on manually-annotated data}
\label{tab:quote-extraction}
\end{table}
\vspace{-0.2cm}

The evaluation of speaker identification when taken independently (Table \ref{tab:speaker-eval}) shows that almost 4 speakers out of 5 are correctly identified. The missing speakers, however, will impact the next step. Although the verbs are not used anymore downstream in the pipeline, their identification is encouraging and could be useful for other applications.

\vspace{-0.2cm}
\begin{table}[ht]
\centering
\begin{tabular}{lccc}
\midrule
                & \cellcolor[HTML]{EFEFEF}\textbf{Precision} & \cellcolor[HTML]{EFEFEF}\textbf{Recall} & \cellcolor[HTML]{EFEFEF}\textbf{F1-score} \\ \midrule
\cellcolor[HTML]{EFEFEF}Speakers (independently) & 79.1\%                                     & 78.8\%                                  & 79.0\%                                    \\
\cellcolor[HTML]{EFEFEF}Verbs (independently)    & 84.5\%                                     & 83.8\%                                  & 84.1\%                                    \\ \bottomrule
\end{tabular}
\caption{Speaker and verb evaluation}
\label{tab:speaker-eval}
\end{table}
\vspace{-0.2cm}
% \vspace{-0.2cm}
% \subsection{Recognizing people}
% \vspace{-0.2cm}

% Identifying all the people in the text and their mentions is essentially a coreference resolution task specifically for Person Type entities. The choice of metrics and the calculation for this type of task is often a challenge itself. Besides, to be able to compute those metrics, we need a corpus annotated with all mentions of person entities. Currently we only possess a corpus annotated with the mentions of speakers (which is only a subset of the mentions of all the named people referred to in the text). For those reasons, the entity merger has no clear evaluation metrics for now.

\vspace{-0.2cm}
\subsection{Mapping speakers of quotes to their references}
\vspace{-0.2cm}

Mapping the speakers of the quotes to their respective most representative person named entity is a crucial task that we call Quote merging. Quote merging can be constructed as an open classification task (i.e., with an unlimited number of classes). Given a set of speakers, the purpose of the quote merger is to assign to each speaker its correct corresponding reference, i.e., the most representative mention of that entity in the same document.

The evaluation of the quote merger considers a corpus of documents annotated with their speakers (\texttt{speakers-GOLD}) and corresponding references (\texttt{references-GOLD}). \texttt{speakers-GOLD} is given as input to the quote merger and the system will predict \texttt{references-SYS}. We then compare \texttt{references-SYS} and \texttt{references-GOLD} to obtain the performance scores as follows: A reference for a given speaker is considered correct (and thus counted as a true positive) if after lowercasing, the Levenshtein distance between \texttt{references-GOLD} and \texttt{references-SYS} is less than 2 (so as to take typos into account). More simply, it can be said that a reference is counted as correct if for a given \texttt{speakers-GOLD}:

\vspace{-0.2cm}
\begin{equation}
    referenceGOLD ~= referenceSYS
\end{equation}

The correct references (\textit{CorrectReferences}) are counted across all documents. In conjunction with this, we also count the total number of references given by the system (\textit{SystemReferenceCount}) and the total number of references in the GOLD annotation (\textit{GOLDReferenceCount}). Those two numbers may differ since the system does not necessarily assign a reference to each speaker. In the end we compute the following metrics:

\vspace{-0.2cm}
\begin{equation}
    Recall = \frac{CorrectReferences} {GOLDReferenceCount}
\end{equation}

\vspace{-0.2cm}
\begin{equation}
    Precision = \frac{CorrectReferences} {SystemReferenceCount}
\end{equation}

%\vspace{-0.2cm}
\begin{equation}
   F1 = 2 \times \frac{(Precision \times Recall)} {(Precision + Recall)}
\end{equation}

The recall and precision are an average of all the speaker/reference pairs \textit{across} documents, not an average of the speaker/reference pairs \textit{per} document. Since each speaker/ reference pair is weighted equally in the final score, documents containing more speakers contribute more to the final score than documents with few speakers.

It is important to note that, since the quote merger relies heavily on the output of the preceding step, the evaluation of the quote merger can also be considered as a good insight into the performance of the named entity recognition stage. 

We reach good precision with our quote merger: a large majority of assigned references are correct, as seen in Table \ref{tab:speaker-ref}. The few precision errors come from coreference errors where pronouns were wrongly assigned to other people of the same gender. Since personal pronouns in French are often gender dependent, such errors do not impact the gender ratio.

\vspace{-0.2cm}
\begin{table}[ht]
\centering
\begin{tabular}{lccc}
\midrule
                & \cellcolor[HTML]{EFEFEF}\textbf{Precision} & \cellcolor[HTML]{EFEFEF}\textbf{Recall} & \cellcolor[HTML]{EFEFEF}\textbf{F1-score} \\ \midrule
\cellcolor[HTML]{EFEFEF}Speaker reference & 90.3\% & 66.6\% & 76.7\% \\ \bottomrule
\end{tabular}
\caption{Speaker reference evaluation}
\label{tab:speaker-ref}
\end{table}
\vspace{-0.2cm}

The recall, although lower than the precision, is also acceptable for our purposes: Less than 25\% of the speakers have a missing reference. Based on visual inspection of the results, there are several cases that explain the misses: 

\begin{small}
\begin{itemize}
    \item A speaker pronoun is wrongly assigned a non-person reference during the coreference resolution. This can happen when pronouns have contexts that are compatible with different types of entities, for instance, between those of type `Person' and `Organization' (because organizations are often personified in speech).
    \item The speaker is further away from the preceding coreferring mention than is allowed by \texttt{coreferee}. In our implementation of the coreference algorithm for French, we specify a `distance limit' (5 sentences for pronouns and 3 sentences for nouns), beyond which the chains will not necessarily be merged. Thus, some chains containing speakers do not contain named speakers, because the speaker was named earlier than the distance limit. 
    %While the specified limit is generally large enough to cover cases of coreference in one-voiced press text, the presence of quote segments inside the texts can cause the mentions of the speakers to fall on each side of the quote and thus the distance between them to exceed the limit.
    \item Some rarer names are never identified as person-type named entities by \texttt{spaCy}’s NER model. As a consequence, they are not included in the list of potential references and can never be associated with a speaker.
\end{itemize}
\end{small}

\vspace{-0.2cm}
\subsection{Gender annotation}
\vspace{-0.2cm}

\subsubsection{Entire pipeline}
\vspace{-0.2cm}

This step measures how many of the people mentioned in the text were correctly detected by our system and how many were missed. According to the human annotation instructions, the most complete name of each person in the text needs to be provided by the annotators in the annotation files.

Using these manually-annotated sets, we can calculate the number of entities our system detects and misses. We first convert all system- and expert-annotated entities in these lists to lowercase and trim the start/end space characters. We consider the human annotated sets ($true\_names$) and the sets produced by the system ($pred\_names$) and perform set operations to obtain:

\begin{itemize}
    \item $True Positives : true\_names \cap pred\_names$
    \item $False Positives: pred\_names / true\_names$
    \item $False Negatives: true\_names / pred\_names$
\end{itemize}

Then we sum up those over all the articles to calculate the precision, recall, and F1-score of each identification task. The evaluation of the entire pipeline helps us to assess the performance of all the previously evaluated components when put together. Evaluating all components of the pipeline independently also allows us to assess the impact of each part of the pipeline on the resulting gender annotation.

As we see in Table \ref{tab:pipeline-eval}, we reach an excellent F1 for people mentioned and thus an excellent F1 for women mentioned and men mentioned (which is expected given the high performance of gender prediction). Precision is slightly better than people which reflects the performance of both the named entity recognizer and the boundary fixes performed afterwards: almost all named entities recognized as people are indeed people, and few people are missed. This also reflects the quality of entity unification: Nearly all people appear to be correctly merged. Note that there were no non-binary or unknown people in the evaluation set. 

The performance for identification of sources is foreshadowed by the numbers obtained when evaluating quote merging. Precision is much better than recall and the latter is still reliable enough to detect the majority of speakers. Interestingly enough, both precision and recall for sources are higher than the multiplication of the metrics for quote merging and speaker extraction. This shows that some errors in the speaker extraction and quote merging can actually be helpful, by catching a source that was not identified by the correctly identified speakers and references. 
For instance, a speaker \textit{il} can be incorrectly marked as a speaker and then said speaker is correctly mapped to the reference \textit{Tom Lanneau}. \textit{Tom Lanneau} is actually a reference to the speaker \textit{le professeur} elsewhere in the text, but the coreference algorithm did not correctly resolve coreference for this occurrence. As a consequence, \textit{Tom Lanneau} would be correctly included in the sources by accident.

\vspace{-0.2cm}
\begin{table}[ht]
\centering
\begin{tabular}{cccc}
\midrule
\multicolumn{1}{l}{}                    & \cellcolor[HTML]{EFEFEF}\textbf{Precision} & \cellcolor[HTML]{EFEFEF}\textbf{Recall} & \cellcolor[HTML]{EFEFEF}\textbf{F1-score} \\ \midrule
\cellcolor[HTML]{EFEFEF}People mentioned         & 96.4\%                                    & 88.7\%                                    & 92.4\%                                    \\
\cellcolor[HTML]{EFEFEF}Women mentioned   & 89.9\%                                    & 84.9\%                                 & 87.3\%                                   \\
\cellcolor[HTML]{EFEFEF}Men mentioned     & 96.7\%                                    & 88.1\%                                 & 92.2\%                                   \\
\cellcolor[HTML]{EFEFEF}Unknown People  & N/A                                        & N/A                                     & N/A                                       \\
\cellcolor[HTML]{EFEFEF}Sources         & 93.9\%                                    & 57.1\%                                 & 71.0\%                                   \\
\cellcolor[HTML]{EFEFEF}Women Sources  & 83.3                                       & 50.0\%                                  & 62.5\%                                   \\
\cellcolor[HTML]{EFEFEF}Men Sources    & 96.0\%                                    & 58.7\%                                 & 72.8\%                                   \\
\cellcolor[HTML]{EFEFEF}Unknown Sources & N/A                                        & N/A                                     & N/A    \\    \bottomrule     
\end{tabular}
\caption{Evaluation of gender prediction in the entire pipeline}
\label{tab:pipeline-eval}
\end{table}
\vspace{-0.2cm}

\vspace{-0.2cm}
\subsubsection{Gender ratio}
\vspace{-0.2cm}

We compared the gender ratio, which is the ratio of women and men sources predicted by our system to those of our human annotations, with results in Table \ref{tab:gender-ratio}. We aim for the system ratio to be the closest possible to the actual (manually-annotated) ratio. The ratios for both the people mentioned and sources quoted are very close to the actual ratios. Most of the classification errors are explained by the algorithm wrongly categorizing women mentioned as men. This is a common problem in French, because some names are associated with a different gender in English and French, and our system uses the same lookup table for English and French. An example is the first name \textit{Jean}, typically female in English, but typically male in French, as \textit{Jéan}, potentially explaining why our system overshoots the human annotations for the gender ratio of women mentioned. 
%It is also expected that some very rare first names are wrongly classified. 
Overall, we observe that our gender prediction algorithm is able to deliver rather accurate results.

Judging by these metrics, any improvement to this stage should come from speaker identification (part of quote extraction) and coreference resolution (part of quote merging).

\vspace{-0.2cm}
\begin{table}[ht]
\centering
\begin{tabular}{cccc}
\toprule
\multicolumn{1}{l}{}                     & \cellcolor[HTML]{EFEFEF}\textbf{Men} & \cellcolor[HTML]{EFEFEF}\textbf{Women} & \cellcolor[HTML]{EFEFEF}\textbf{Unknown/Other} \\ \midrule
\multicolumn{4}{c}{\textbf{People}}        \\ \midrule
\cellcolor[HTML]{EFEFEF}Human annotation & 73.5\%                                & 26.5\%                                  & 0.0\%                                    \\
\cellcolor[HTML]{EFEFEF}System           & 72.7\%                                & 27.3\%                                  & 0.0\%                                    \\ \midrule
\multicolumn{4}{c}{\textbf{Sources}}        \\ \midrule
\cellcolor[HTML]{EFEFEF}Human annotation & 75.3\%                                & 24.7\%                                  & 0.0\%                                    \\
\cellcolor[HTML]{EFEFEF}System           & 75.5\%                                & 24.5\%                                  & 0.0\%    \\   \bottomrule        
\end{tabular}
\caption{Evaluation of gender ratio}
\label{tab:gender-ratio}
\end{table}
\vspace{-0.2cm}

%\vspace{-0.2cm}
\section{The gender representation picture in French Canadian media}
\vspace{-0.2cm}

We analyzed 15 months of news stories in French Canadian media, between October 1, 2021 and December 31, 2022, for a total of 282,512 news stories. The overall gender breakdown in these news articles is: 71.5\% men quoted, 28.3\% women quoted, and 0.2\% unknown or non-binary people quoted. Table \ref{tab:outlets} shows a breakdown of these numbers per news organization. 

\vspace{-0.2cm}
\begin{table}[ht]
\centering
\begin{tabular}{lcccc}
\toprule
\textbf{Organization}               & \textbf{\% Men} & \textbf{\% Women} & \textbf{\% Unknown/Other} & \textbf{Total articles} \\ \midrule
\rowcolor[HTML]{EFEFEF} 
Le Journal de Montréal & 74.6\% & 25.2\% & 0.2\% & 60,300 \\
La Presse & 72.2\% & 27.6\% & 0.2\% & 61,674 \\
\rowcolor[HTML]{EFEFEF} 
Le Devoir & 71.6\% & 28.2\% & 0.1\% & 35,706 \\
Le Droit & 71.8\% & 28.1\% & 0.1\% & 31,623 \\
\rowcolor[HTML]{EFEFEF} 
Radio Canada & 66.9\% & 32.9\% & 0.1\% & 65,280 \\
TVA Nouvelles & 71.8\% & 28.0\% & 0.2\% & 27,929 \\ \midrule
\textbf{Total} & 71.5\% & 28.3\% & 0.2\% & 282,512\\ \bottomrule
\end{tabular}
\caption{Percentages of sources per outlet, October 1, 2021 to December 31, 2022}
\label{tab:outlets}
\end{table}
\vspace{0.2cm}

We also wanted to explore in which capacity the most frequent sources are quoted, i.e., whether we hear more from politicians, celebrities, or experts. We extracted the 100 top men and women sources for each of the 15 months we studied, and manually annotated the profession of each of those sources. We arrived at a list of top professions or categories, based on existing work \cite{Morris16-GOS}, as the most typical categories of those quoted in the news. The results are presented in Table \ref{tab:top-sources}. 

There are some interesting observations we can derive from the results. The most quoted persons are politicians, for both men and women. \textit{Sports} and \textit{Unelected government officials} are second and third for men, and third and second for women. A significant difference in gender is in \textit{Health profession}, where a much higher percentage of women are quoted. There are more errors or unknowns for women, for two reasons. Many names of men end up being categorized as women. The second reason is that names are not unique enough to help us decide why the person was quoted (and are thus classified as \textit{Unknown}). When we search for \textit{Isabelle Côte} or \textit{Anna Walker}, it is unclear which individual those names refer to, since there may be several people with those names in the news.

\vspace{-0.2cm}
\begin{table}[ht]
\centering
\begin{tabular}{lrrrr}
\toprule

                                   & \multicolumn{1}{c}{\textbf{Men}}          & \multicolumn{1}
                                   {c}{\textbf{}}   & \multicolumn{1}{c}{\textbf{Women}}        & \multicolumn{1}{c}{\textbf{}}   \\ \midrule
\textbf{Profession}                & \multicolumn{1}{c}{\textbf{\# of quotes}} & \multicolumn{1}{c}{\textbf{\%}} & \multicolumn{1}{c}{\textbf{\# of quotes}} & \multicolumn{1}{c}{\textbf{\%}} \\ \midrule
\rowcolor[HTML]{EFEFEF} 
Politician                         & 160,581                                   & 72.3\%                          & 44,059                                    & 47.7\%                          \\
Sports                             & 19,071                                    & 8.6\%                           & 2,611                                     & 2.8\%                           \\
\rowcolor[HTML]{EFEFEF} 
Unelected   govt. official         & 11,795                                    & 5.3\%                           & 8,049                                     & 8.7\%                           \\
Health profession                  & 13,205                                    & 5.9\%                           & 21,553                                    & 23.4\%                          \\
\rowcolor[HTML]{EFEFEF} 
Leader   (union, school, activist) & 3,447                                     & 1.6\%                           & 2,207                                     & 2.4\%                           \\
Police                             & 2,201                                     & 1.0\%                           & 2,031                                     & 2.2\%                           \\
\rowcolor[HTML]{EFEFEF} 
Private   business                 & 2,124                                     & 1.0\%                           & 2,144                                     & 2.3\%                           \\
Legal profession                   & 2,251                                     & 1.0\%                           & 1,620                                     & 1.8\%                           \\
\rowcolor[HTML]{EFEFEF} 
Creative   industries              & 2,479                                     & 1.1\%                           & 1,610                                     & 1.7\%                           \\
Perpetrator                        & 1,273                                     & 0.6\%                           & 360                                       & 0.4\%                           \\
\rowcolor[HTML]{EFEFEF} 
Academic/researcher                & 1,602                                     & 0.7\%                           & 1,716                                     & 1.9\%                           \\
Victim/witness                     & 869                                       & 0.4\%                           & 1,775                                     & 1.9\%                           \\
\rowcolor[HTML]{EFEFEF} 
Media                              & 628                                       & 0.3\%                           & 427                                       & 0.5\%                           \\
Non-govt.   organization           & 430                                       & 0.2\%                           & 1,186                                     & 1.3\%                           \\
\rowcolor[HTML]{EFEFEF} 
Error/unknown                      & 300                                       & 0.1\%                           & 728                                       & 0.8\%                           \\
Person on the street interviews    & 0                                         & 0.0\%                           & 225                                       & 0.2\%                           \\ \midrule
\textbf{Total}                     & \textbf{222,256}                          & \textbf{}                       & \textbf{92,301}                           & \textbf{}                       \\ \bottomrule
\end{tabular}
\caption{Top 100 men and women sources, by category, in each of the 15 months  between October 1, 2021 and December 31, 2022}
\label{tab:top-sources}
\end{table}
%\vspace{-0.2cm}

\vspace{0.2cm}
\section{Discussion and conclusions}
\vspace{-0.2cm}

We present an NLP system  based on linguistic-motivated heuristics, built on top of a robust open-source framework, \texttt{spaCy}. Our contributions can be summarized as: First, we defend the need for NLP systems in applied contexts that rely on linguistic analyses. Second, we contribute a robust coreference resolution methodology for French text, thoroughly tested and deployed in a live environment. 

The success of large language models and end-to-end-systems in NLP seems to have relegated linguistic-based NLP to a second plane. In contexts where it is impossible to annotate enough data to reach acceptable accuracy, and where a pipeline approach seems reasonably accurate, we advocate for the latter. It is concerning, however, that recent research advancing progress in NLP (POS tagging, parsing, coreference resolution, discourse parsing, semantic analysis) is diminishing in its generality and that flexible, open-source systems that generalize across languages are not widely available for languages beyond English. Our work shows that linguistics-driven approaches can be successful, robust, and produce reliable results at scale. As Richard Sproat suggests, \cite{Sproat22-BPA}, sometimes `boring' problems require traditional solutions. 

We also contributed to applied NLP by extending the \texttt{coreferee} library to support French, a capability that did not exist before we began this work.

A final contribution is our visualization and dissemination of the statistics of gender representation in French media to the general public, through the Radar de Parité. Our results and analyses show a persistent underrepresentation of women and a practical absence of non-binary people in Canadian French language media. 

\vspace{-0.2cm}
\section*{Acknowledgements}
\vspace{-0.2cm}
{\small The Radar de Parité and its English counterpart, the Gender Gap Tracker, are a large team effort. We thank especially members of the Research Computing Group at Simon Fraser University, present and past, who maintain the database and support data analyses: Jillian Anderson, Philip Chen, Alexandre Lopes. This research was initiated and supported by \href{https://informedopinions.org/}{Informed Opinions}, a non-profit dedicated to amplifying the voices of women and gender-diverse people. Funding support: Informed Opinions, Simon Fraser University, Social Sciences and Humanities Research Council, Natural Sciences and Engineering Research Council. }

\begin{spacing}{0.2}
\printbibliography[heading=subbibintoc]
\end{spacing}

\newpage

\appendix
\section{Named Entity Recognition steps}
\label{appendix:ner}
\vspace{-0.2cm}

\subsection{Mention head identification}
\label{sec:mention-head}
\vspace{-0.2cm}

To detect mentions, we use a \texttt{spaCy}-based utility called \textit{Rules Analyzer}. This is a component of the coreference resolution library we used for this project, \texttt{coreferee}.\footnote{\url{https://github.com/explosion/coreferee}} The analyzer can also be used independently from \texttt{coreferee} for other tasks, in this case, for identifying mention heads.

In coreference resolution, a mention is linguistically either a noun phrase or a pronoun that can potentially corefer, i.e., refer to another entity. This definition of mention is also compatible with named entities since they are generally noun phrases. A mention can thus be defined as a list of tokens containing a noun/pronoun. For example, in Example (\ref{ex:the-president}), there are five mentions, listed below the text.

\begin{small}
\begin{exe} 
\ex\label{ex:the-president}
\textit{Le Président de l’Entreprise de Pétrole et sa femme ont voyagé à Delhi.} \\
`The President of the Oil Company and his wife traveled to Delhi.' \\
\textbf{Mentions}: \\ 
a) \texttt{The President of the Oil Company} \\
b) \texttt{The Oil Company }\\
c) \texttt{His wife} \\
d) \texttt{Delhi} \\
e) \texttt{The President of the Oil Company and his wife }(since they together form a plural real world entity that can corefer on its own)
\end{exe}
\end{small}

In our definition of mention, however, several problems arise when dealing with the task of aligning between non-identical mentions. Given two identified mentions (e.g., `The President' and `Oil Company'), both mentions would be considered equally valid/invalid identifications of mention \textbf{a}. This is intuitively unsatisfactory, as `President' is the head of the noun phrase and, as animate and human, more salient. From a pragmatic standpoint, being able to identify \textbf{a} would generally be considered more useful than identifying \textbf{b}. The obstacle is that mention \textbf{b} is embedded in mention \textbf{a}. As a consequence, any attempt to align \textbf{b} based on token overlap would also align \textbf{a}, making the two mentions indistinguishable.
Similarly, coordinated noun phrases such as \textbf{e} should also be able to be distinguishable from the noun phrases that form them (\textbf{a} and \textbf{c}). 

This is why we consider neither the whole list of tokens, nor individual tokens to define a mention, but, rather, the heads of that mention. In the example above, the mentions considering that new definition would be:

\begin{small}
\begin{exe}
\ex a) \texttt{President} \\
b) \texttt{Company} \\
c) \texttt{wife} \\
d) \texttt{Delhi} \\
e) \texttt{{[President, wife]}}
\end{exe}
\end{small}

This approach resolves the three issues by boiling down the mentions to one or a few representative tokens (the heads). Note that in the case of coordination, the mention can be composed of several heads which are sibling tokens connected by a link of coordination (as seen with \textbf{e}). Those heads are proper/common nouns or pronouns.

The Rules Analyzer (more specifically its French version) uses \texttt{spaCy}’s dependency parse tree and PoS-tagging to output the lists of mention heads corresponding to the more traditional mention noun phrases and to output all the coordinated siblings of a given mention head.

\subsection{Named Entity Recognition}
\begin{figure}[h]
    \centering
    \includegraphics[scale=0.4]{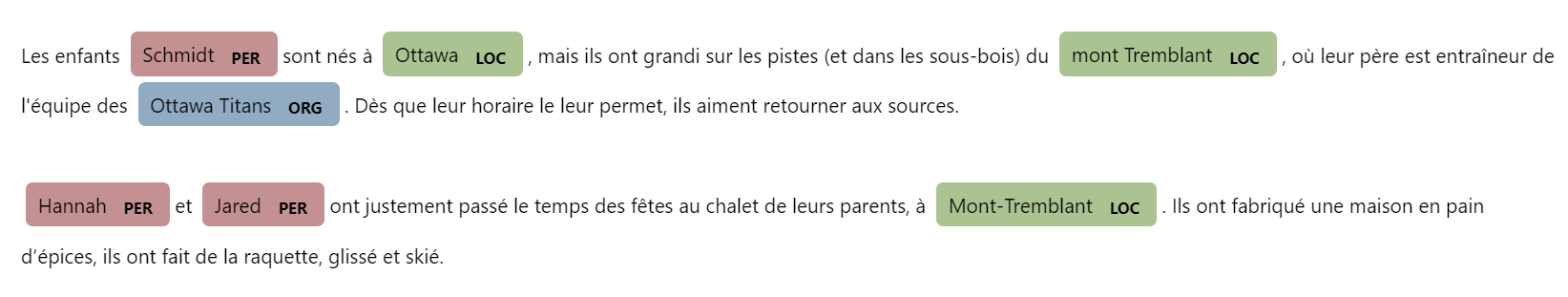} \\
    \caption{Output of \texttt{spaCy}'s NER on a sample text}
    \label{fig:ner}
\end{figure}

To find the people mentioned in a given text, we extract named entities using the existing off-the-shelf \texttt{spaCy} French language model. We only keep named entities of the type PER (person). In Figure \ref{fig:ner}, an example text is shown and the extracted named entities are highlighted. For clarity, just the PER, ORG (organization) and LOC (location) tags are displayed. The model quite reliably distinguishes between entities that are people and others that represent places or organizations --- this is important for downstream tasks such as attributing quotes to a human speaker.

\vspace{-0.2cm}
\vspace{-0.2cm}

Generally, the algorithm does a good job at finding persons in the text. In order to tackle recurring errors, several rules were implemented on top of the named entity recognition performed by \texttt{spaCy}. We list some of these recurring problems below, as well as the methods we used to deal with them.

\vspace{-0.2cm}
\subsubsection{Entity misclassification}
\vspace{-0.2cm}

There are cases of recurring entity names that are consistently mislabelled by \texttt{spaCy}. Although they are correctly identified as named entities, the label assigned to the entity is incorrect. There are two possible cases:

\begin{small}
\begin{itemize}
\item An entity referring to non-people (LOC or ORG) is labelled as PER. In that case, the entity would be incorrectly included in the list of people and this would create noise.
\item Conversely, some names of people are labeled as LOC, ORG, or MISC. This can be caused by nouns being both people names and location names (e.g., ``Caroline''). As a consequence, those entities would be missing from the final list of people.
\end{itemize}
\end{small}

We added an extra component to \texttt{spaCy}’s Named Entity pipeline component called the Entity Ruler. This component allows us to change the default label of the entities given by \texttt{spaCy} by supplementing it with a list of patterns (typically entity text) mapped to the correct entity. We identified the most frequent cases of misclassification in the news data and added their patterns and corresponding labels to the list.

\vspace{-0.2cm}
\subsubsection{Incorrect entity boundaries}
\vspace{-0.2cm}

\begin{figure}[h]
    \centering
    \includegraphics[scale=0.4]{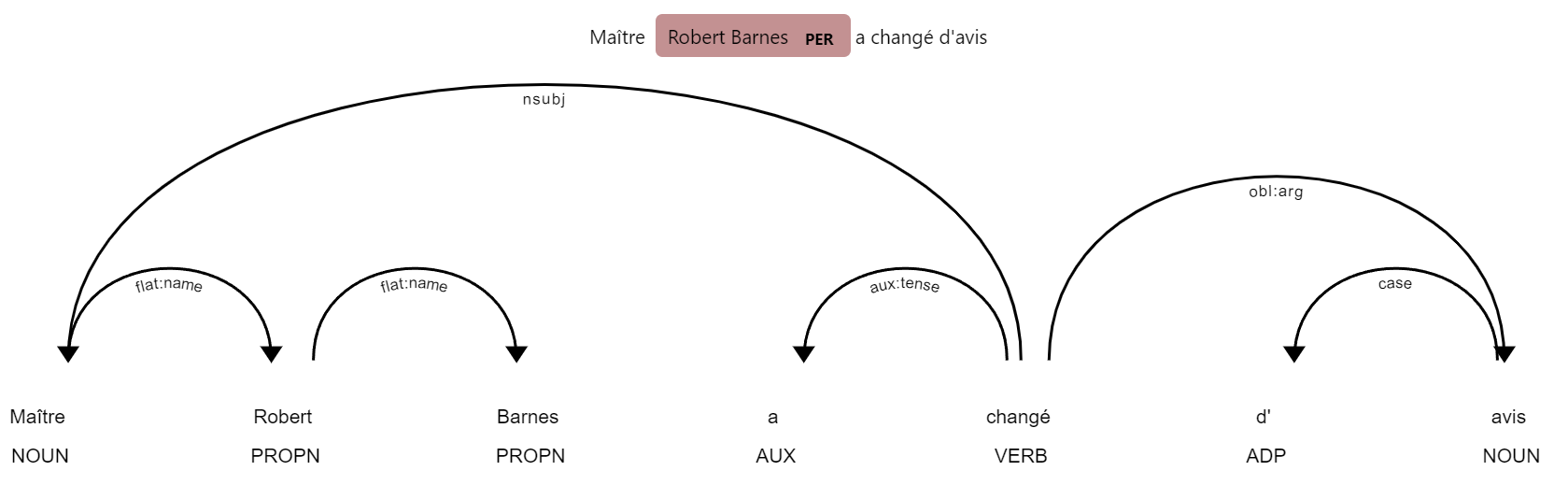} \\
    \caption{Wrong NER boundary (top) and dependency parse for the sentence}
    \label{fig:ner-error}
\end{figure}

Since the named entity recognition is independent from \texttt{spaCy}’s parse tree, the boundaries of the named entity extracted by \texttt{spaCy} do not necessarily correspond to those of a noun phrase within the same text. There are two possible instances of this problem: 

\begin{small}
\begin{itemize}
    \item The beginning of the entity is typically missing with some titles such as \textit{Docteur} or \textit{Maître}. In Figure \ref{fig:ner-error}, we see the output of the NER at the top, with the PER label applied only to \textit{Robert Barnes} and not to the entire name, \textit{Maître Robert Barnes}. 
    In those cases we extend the entities to the left of the name token by a dependency relation of type ``flat:name'' (the dependency in \texttt{spaCy} that links different parts of a name; see bottom of Figure \ref{fig:ner-error}). 
    Since the most common person titles (\textit{Monsieur, Madame}) are generally included in the entity in our French article dataset, this step allows us to normalize the patterns for the entities that follow.

    \item In cases with long names (more than three tokens), the end of the entity typically gets missed. We use the same method as above to extend the entity to the right, using the right side of the entity as the anchor.
\end{itemize}
\end{small}

%\textcolor{red}{TO DO: improve the figure (better resolution). Also, shouldn't `Maître' have a circumflex? }

\vspace{-0.2cm}
\subsubsection{Hyphenated names}
\vspace{-0.2cm}

\begin{figure}[h]
    \centering
    \includegraphics[scale=0.9]{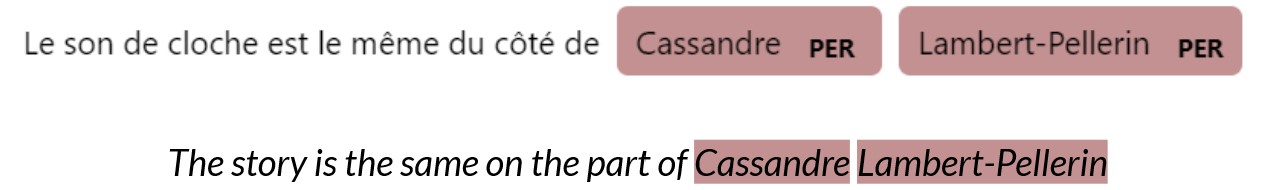} \\
    \caption{Wrong NER boundary for a hyphenated name}
    \label{fig:ner-error2}
\end{figure}

Hyphenated names are common in French (\textit{Jean-Michel, Marie-Josée}). However \texttt{spaCy}’s tokenization approach tends to capture hyphens as separate tokens, which can cause problems for us during gender identification, as we see in Figure \ref{fig:ner-error2}, where the same person is identified as two different entities. The approach described earlier for locating entity boundaries is not reliable enough to handle hyphens, since they are often considered sentence boundaries, and can lead to very inconsistent parsing in French. 

For this reason, we ignore the parse tree when dealing with hyphens in French, and, instead simply look for hyphens following the end of entities (either immediately or one token beyond). We go on the assumption that all the tokens belonging to the named entity are capitalized and we extend the entity after the hyphen until the next non-capitalized token. Through this method, we are able to extend the named entity to capture names such as \textit{Cassandre Lambert-Pellerin} from Figure \ref{fig:ner-error2}.

\vspace{-0.2cm}
\vspace{-0.2cm}

\vspace{-0.2cm}
\subsubsection{Coordinated siblings}
\vspace{-0.2cm}

Named entities made up of more than one sibling are sometimes parsed erroneously. For Example (\ref{ex:pierre-dupont}), only the individuals are identified as entities, not the coordinated entity \textit{Pierre Dupont et Marie Jugneau}.

\begin{small}
\begin{exe}
\ex\label{ex:pierre-dupont} \textit{Pierre Dupont et Marie Jugneau mangent. Gérard Klein et sa famille les rejoignent.}\\
    `Pierre Dupont and Marie Jugneau are eating. Gérard Klein and his family join them.' \\
    \textbf{Entities}: \\
    a) \texttt{Pierre Dupont} \\
    b) \texttt{Marie Jugneau} \\
    c) \texttt{Gérard Klein}
\end{exe}
\end{small}

We make use of the Rules Analyzer to check whether the coordinated siblings of the head of each entity correspond to subsequent named entities. Here, we find that the sibling of the head of entity \textbf{a}, \textit{Pierre} (i.e., \textit{Marie}) is also the head of entity \textbf{b} and we would consequently add the new entity they form together entity \textbf{d}, \textit{Pierre Dupont et Marie Jugneau}. When the coordinated siblings of a named entity do not match with any entity, we simply add it as a new entity. We hypothesize that noun phrases in a coordination-type dependency with a PER named entity are always people (even though there may be few exceptions, such as \textit{Jack and the company}). Entity \textbf{c} is in such a case and thus would prompt us to add the new entity \textbf{e}, \textit{Gérard Klein et sa famille}.

\vspace{-0.2cm}

\subsubsection{Nobility titles}
\vspace{-0.2cm}

Some last names in French include current or vestigial nobility titles. These titles are typically introduced by the preposition \textit{de}. However, as shown in Example (\ref{ex:anne-marie}), there are cases of prepositional phrases introduced by \textit{de} that follow a named entity but are not part of the name --- those phrases are often included in the named entity by \texttt{spaCy}.

\begin{small}
\begin{exe}
\ex\label{ex:anne-marie} \textit{ [\ldots] un conflit s’y déroule et une personne saigne à la tête, a-t-elle rapporté à la juge Anne-Marie Jacques de la Cour du Québec qui préside le procès. }\\
      `a conflict is taking place there and a person is bleeding from the head, she reported to judge Anne-Marie Jacques of the Court of Quebec who is presiding over the trial.'
\end{exe}
\end{small}

In order to keep only the nobility titles and not the other adpositional phrases, we implemented an algorithm to trim the entities containing a preposition that follows several conditions:

\begin{small}
\begin{itemize}
    \item any adpositional phrase that is not introduced by \textit{de} is trimmed.
    \item any adpositional phrase containing a common noun, whether capitalized or not is trimmed
    \item any adpositional that does NOT contain a name is trimmed. Names are identified by the following conditions
    \begin{itemize}
        \item being a proper noun
        \item being capitalized
        \item not being in uppercase
        \item not being a mail address
        \item not being an url
    \end{itemize}
\end{itemize}
\end{small}

Using this method on the above example we are able to fix the boundaries of the named entities to obtain the correct entity \textit{Anne-Marie Jacques}. 

\vspace{-0.2cm}
\subsubsection{Entity trimming}
\vspace{-0.2cm}

Lastly, entities recognized by \texttt{spaCy} can be overly long and span over punctuations and even line breaks. Since in French texts, people’s names can only be made of letters and hyphens, we truncate the entities left and right starting from any characters that are neither a hyphen, nor a letter, nor a whitespace:

\begin{small}
\begin{exe}
\ex \label{ex:julien-jean}
    \textit{Julien Jean} \\
    \textit{Dupont est arrivé} \\
    'Julien Jean // Dupont has arrived.'
\end{exe}
\end{small}

Here the two parts of the entities are separated by a newline. We would trim the right side of the entity to obtain only \textit{Julien Jean}.

\section{Extracting quotes} \label{appendix:quote-extraction}
\vspace{-0.2cm}

\subsection{Direct quotes}
\label{sec:direct-quotes}
\vspace{-0.2cm}

Direct quotes are identified by searching for an opening quotation mark and keeping the quote open until the corresponding closing mark is found (this includes multi-level quotation marks).

A special type of direct quotes are what we have labelled as floating quotes. When multiple quotes by the same speaker are present in a news article, it is often the case that only one quotative structure is used (direct or indirect), with subsequent quotes receiving their own sentence or sentences, with the entire sentence within quotes, as in the example below, where the last quoted portion is to be interpreted as having been said by Enrico Ciccone, the last speaker of an indirect quote.

%https://www.tvanouvelles.ca/2019/02/08/un-carnet-des-commotions-cerebrales-pour-les-jeunes-1
\begin{small}
\begin{exe}
\ex    \textit{Le député libéral Enrico Ciccone propose de créer un dossier qui permettra de suivre à la trace les antécédents de commotion cérébrale des jeunes tout au long de leur parcours sportif. [\ldots] «Je veux protéger les enfants avant qu’ils arrivent à la majorité, affirme-t-il, [\ldots]». «Quand je vais crever et qu’ils vont ouvrir mon cerveau, vont-ils voir de l’encéphalopathie? Je vais-tu faire de la démence dans dix ans? Je vais-tu faire de l’Alzheimer dans dix ans? Tous les matins, je me lève avec cette inquiétude-là.». } 

Liberal MP Enrico Ciccone proposes to create a record that will track the concussion history of young people throughout their athletic journey. [\ldots] ``I want to protect children before they come of age, he says, [\ldots].'' ``When I die and they open my brain, will they see encephalopathy? Will I have dementia in ten years? Will I have Alzheimer's in ten years? Every morning, I get up with this worry.''
\end{exe}
\end{small}

Detecting floating quotes with regular expressions (regex) is relatively easy and accurate. Unfortunately, it often happens that the speaker of a floating quote is not mentioned in the sentence immediately preceding that second quote, but instead in some sentence beforehand. Additionally, it is not strictly necessary for the speaker to have been quoted before---sometimes a speaker might have been mentioned in a previous sentence without a quote. 

Finally, there is another type of floating quote common in French texts that does not exist in English: \textit{la phrase incise}. This type of quote embeds the verb and speaker between the quotation marks themselves, such in the example below. The subject and verb of the direct speech, \textit{dit-il} are enclosed within the quotation marks. This type of quote is by far the hardest to predict, as there may be a quote indicating verb in the quote that may or may not be an actual quote indicating verb for that quote.

%https://plus.lapresse.ca/screens/4130e87a-3488-40ed-8b85-946bcc2064eb__7C___0.html
\begin{small}
\begin{exe}
\ex 
    \textit{Cet acteur, c’est Guillaume Cyr. Je lui ai parlé cette semaine. Il n’avait pas décoléré. «On est très tolérants, dit-il. On les laisse texter, on les laisse rire. Mais il y a des choses qui sont inacceptables. On se donne pour eux, de la même manière qu’on le ferait pour des gens qui paient leur billet 80\$ le vendredi soir. On est là devant eux. Ce n’est pas du cinéma!»} 

    This actor is Guillaume Cyr. I spoke to him this week. He had not taken off. ``We are very tolerant, he said. We let them text, we let them laugh. But there are things that are unacceptable. We give ourselves for them, in the same way as we would for people who pay their \$80 ticket on Friday evening. We are there in front of them. This is not cinema!''
\end{exe}
\end{small}

In general, however, direct quotes are easier to identify, as the typography aids in marking the beginning and end. Existing quote identification systems tend to focus on direct quotes only \cite{Balahur09-OMO,Morris21-TSU}. 

\subsection{Indirect quotes}
\label{sec:indirect-quotes}
\vspace{-0.2cm}

We define indirect quotes as quotes of one of the following structures in Table \ref{tab:indirect-quotes}.

\begin{table}[h]
\centering
\begin{tabular}{@{}ccc@{}}
\toprule
 \cellcolor[HTML]{EFEFEF}\textbf{speaker}    & \cellcolor[HTML]{EFEFEF}\textbf{quote indicating verb} & \cellcolor[HTML]{EFEFEF}\textbf{quote} \\ 
Elle & dit         & qu'il pleut.    \\
 \cellcolor[HTML]{EFEFEF}\textbf{quote} & \cellcolor[HTML]{EFEFEF}\textbf{quote indicating verb} & \cellcolor[HTML]{EFEFEF}\textbf{speaker} \\ 
Il pleut, & dit                                  & elle.                        \\ \bottomrule
\end{tabular}
\caption{Examples of indirect quotes}
\label{tab:indirect-quotes}
\end{table}

For indirect quotes, we rely on the constituency or syntactic structure extracted for each sentence. We use \texttt{spaCy}’s built-in dependency parser, which is a Convolutional Neural Network (CNN) model, to identify constituency structures in sentences. The current version for French only resembles the English version with regard to the general function structure, with custom methods for capturing language-specific quotative structures in French.

\vspace{-0.2cm}
\subsubsection{Clausal complements}
\vspace{-0.2cm}

The first and main pattern that we are concerned with is captured by the clausal complement structure, represented in sentence parsing with the CCOMP tag \cite{deMarneffe08-TST}. The clausal complement of a verb is a dependent clause with an internal subject, which functions as an object of the verb. For example, in the sentence \textit{He says that you like to swim}, the clause \textit{you like to swim}, whether it contains the complementizer \textit{that} (\textit{que} in French) or not, functions as the clausal complement dependent on the verb \textit{says}. We extract not only the quoted piece of text, but also the reporting verb (\textit{says}) and its subject (\textit{he} in this example). This syntactic structure is, however, very common in both English and French and is not specific to quotations (\textit{He saw that you like to swim; He believes that you like to swim}). To ensure that we are capturing reported speech and not other clausal objects of verbs, we created an `allow-list' of verbs that typically introduce quotes, which are referred to as verbs of communication \cite{levin93}.

\vspace{-0.2cm}
\subsubsection{Quote verb allow-list}
\vspace{-0.2cm}

For the French quote-indicating verb list, we initially started with the list of verbs translated from the English verb allow-list, such that only verbs that actually indicated quotes in French as well were included. We then appended the list with all the verbs from the annotated data that also indicated quotes, which greatly improved performance. Importantly, we finally used \texttt{spaCy}'s French lemma lookup table to vastly increase the size of the quote-indicating verb list, as inverting the dictionary (which stores key-value pairs such that the keys are conjugated forms that map to the infinitive of the verb, which is the value) allowed us to find the many different conjugations for the infinitives already present in the list. The final list of verbs resulted in a much higher performance, even compared to several other modifications to the pipeline. A list of infinitives can be found in Table \ref{tab:quote-verbs} below:

\vspace{-0.2cm}
\begin{table*}[ht]
\centering
\small
\begin{tabular}{p{13.5cm}}
\\\toprule
accuser, admettre, adresser, affirmer, ajouter, alléguer, annoncer, appeler, arguer, assurer, attester, avancer, avertir, avouer, citer, commenter, conclure, concéder, confier, confirmer, considérer, constater, craindre, critiquer, croire, dancer, demander, demeurer, dire, divulger, déclarer, décrire, défendre, dénoncer, déplore, déplorer, désoler, écrire, enchaîner, entendre, espérer, estimer, évoque, exiger, expliquer, exprimer, extasier, gazouiller, illustrer, indiquer, informer, inquiéter, insister, ironiser, juger, justifier, lancer, lire, louanger, mentionner, noter, nuancer, observer, penser, plaider, plaindre, poursuivre, preciser, promettre, proposer, protester, préciser, prédire, prétendre, prétexter, prévenir, prévoir, raconter, rappeler, rapporter, re-connaître, recommander, reconnaitre, reconnaître, redouter, relater, relever, remarquer, remémore, remémorer, réagir, réclamer, réfuter, réjouir, répliquer, répondre, résumer, rétorquer, révéler, saluer, savoir, souhaiter, soulever, souligner, soupire, soupirer, sourire, soutenir, suggérer, tonner, trancher, trouver, tweeter, témoigner, valoir, vouloir
\\
\bottomrule
\end{tabular} 
\caption{Hand-curated allow-list of quote verbs}
\label{tab:quote-verbs}
\end{table*}
\vspace{-0.2cm}

\vspace{-0.2cm}
\subsection{Quotes with \textit{selon}}
\vspace{-0.2cm}

A different syntactic structure involved in quotative contexts features the prepositional phrase \textit{selon} (`according to'). Quotations that use this structure are usually one of two types: following or preceding \textit{selon}. Example (\ref{ex:selon}) shows two examples with regard to this structure.
In the first example, we see a case where \textit{selon} precedes the content of the quote (\textit{la députée ne pourrait\ldots}). The second example shows the construction with the content of the quote at the beginning, before the quotative \textit{selon} and before the name of the speaker of the quote. Note that \textit{selon} quotes may or may not include quotation marks around the actual quotes, as in the second example under (\ref{ex:selon}).
Both of these examples are being captured by the current version of the French quote extractor.

\begin{small}
\begin{exe}
\ex\label{ex:selon}
\begin{xlist}
    \ex \textit{Selon M. Winkler, la députée ne pourrait plus forcer Twitter à prendre des mesures en vertu de l'article 19.17 énoncé dans l'ACEUM.} \\
    `According to Mr. Winkler, the MP could no longer force Twitter to take action under Article 19.17 set out in the CUSMA.' 
    
    \ex \textit{Il s’est dit «très surpris» du jugement du TAQ, «contraire à la jurisprudence rendue», selon lui.} \\
    `He said he was ``very surprised'' by the TAQ's judgment, ``contrary to the case law rendered'', according to him.'
\end{xlist}
\end{exe}
\end{small}

The database record for each extracted quote consists of several fields, including the quote content, its speaker and verb as well as the character indices of each of these text spans within the news article. An example of an extracted quote can be seen in Figure \ref{fig:quote}.

\begin{figure}[ht]
\begin{small}
\begin{minted}[frame=single,
               framesep=2mm,
               xleftmargin=8pt,
               tabsize=3]{json}
{     
    "speaker": "M. Chang",
    "speaker_index" : "(3535, 3543)",
    "quote" : "Des pays latino-américains comme la Colombie et le Chili testent 
    aussie cette technologie", 
    "quote_index": "(3431, 3526)", 
    "verb": "affirme", 
    "verb_index": "(3527, 3534)",
    "quote_token_count": 16, 
    "quote_type": "CVS", 
    "is_floating_quote": false,
    "reference": ""
}
\end{minted}
\caption{Structure of a quote object}
\label{fig:quote}
\end{small}
\end{figure}

\end{document}